\documentclass[journal]{IEEEtran}

\usepackage[T1]{fontenc}
\usepackage{amsmath,amssymb,amsfonts}
\usepackage{algorithmic}
\usepackage{algorithm}
\usepackage{graphicx}
\graphicspath{{figures/}}
\usepackage{textcomp}
\usepackage{xcolor}
\usepackage{booktabs}
\usepackage{multirow}
\usepackage{array}
\usepackage{url}
\usepackage{hyperref}
\usepackage{cite}
\usepackage{subcaption}
\usepackage{siunitx}
\usepackage{balance}
\usepackage{enumitem}

\newcolumntype{L}[1]{>{\raggedright\arraybackslash}p{#1}}
\newcolumntype{C}[1]{>{\centering\arraybackslash}p{#1}}
\newcolumntype{R}[1]{>{\raggedleft\arraybackslash}p{#1}}

\DeclareMathOperator*{\argmax}{arg\,max}

\DeclareMathOperator{\tr}{tr}

\begin{document}

\title{Supervised Dimensionality Reduction Revisited:\\Why LDA on Frozen CNN Features Deserves a Second Look}

\author{
\IEEEauthorblockN{Indar Kumar, Girish Karhana, Sai Krishna Jasti, and Ankit Hemant Lade}
\thanks{Manuscript submitted for review. I.~Kumar, G.~Karhana, S.~K.~Jasti, and A.~H.~Lade are independent researchers (e-mail: indarkarhana@gmail.com, girishkarhana8@gmail.com, jsaikrishna379@gmail.com, ankitlade12@gmail.com).}
}

\maketitle

\begin{abstract}
Frozen pretrained image representations are widely used for transfer learning: a backbone is kept fixed, feature vectors are extracted, and a lightweight classifier is trained on top. This pipeline usually feeds the full feature vector to the classifier, even when the target task has far fewer classes than the pretraining task. We revisit a classical alternative: supervised dimensionality reduction with Linear Discriminant Analysis (LDA) before linear probing.

We evaluate ten dimensionality-reduction strategies on frozen features from six backbones---ResNet-18, ResNet-50, MobileNetV3-Small, EfficientNet-B0, ViT-B/16, and DINOv2-ViT-S/14---across CIFAR-100, Tiny ImageNet, and CUB-200-2011. Under a fixed logistic-regression protocol, LDA improves accuracy over full features in 11 of 12 coarse-grained configurations, with gains up to 4.5 percentage points while reducing feature dimensionality by 48--87\%. The same projection consistently hurts on fine-grained CUB-200, where full features win across all six backbones. This establishes a practical boundary condition: LDA is useful when class-level structure is coarse enough to be captured by mean-separating directions, but it can discard subtle cues needed for fine-grained recognition.

We also compare LDA with PCA, PCA+LDA, regularized LDA, Local Fisher Discriminant Analysis, Neighbourhood Components Analysis, and three lightweight LDA extensions. The results show that plain LDA offers the best accuracy--cost tradeoff for most coarse-grained settings, while more complex supervised reduction methods rarely justify their additional cost. Overall, the study provides concrete guidance for when post-hoc supervised projection should, and should not, be inserted into frozen-feature image classification pipelines.
\end{abstract}


\section{Introduction}
\label{sec:introduction}

The transfer learning paradigm has fundamentally changed how practitioners approach image classification. Rather than training a deep network from scratch on a target task, the standard recipe is to take a network pretrained on ImageNet~\cite{deng2009imagenet}, freeze its weights, extract feature vectors from a late layer, and train a lightweight classifier on top~\cite{donahue2014decaf, sharif2014cnn, kornblith2019better}. This frozen-feature approach is attractive when labeled data is scarce, when computational resources are limited, or when rapid iteration across many target tasks is needed. It applies equally to convolutional networks and to the vision transformers~\cite{dosovitskiy2021vit} and self-supervised models~\cite{oquab2024dinov2} that have recently become prominent.

A frozen ResNet-18, for instance, produces 512-dimensional feature vectors; a ResNet-50 produces 2048-dimensional vectors; a ViT-B/16 produces 768-dimensional vectors. These vectors are then fed to a logistic regression or small MLP. What is rarely questioned is whether the \emph{full} feature vector is the right input to the downstream classifier. After all, these features were optimized for 1000-way ImageNet classification (or, in the case of self-supervised models, for general visual representation). When the target task has far fewer classes---100 in CIFAR-100, 200 in Tiny ImageNet---the features live in a space whose intrinsic dimensionality may be much lower than the ambient one.

Dimensionality reduction before classification is one of the oldest ideas in pattern recognition. Principal Component Analysis (PCA)~\cite{pearson1901pca, jolliffe2002pca} finds directions of maximal variance; Linear Discriminant Analysis (LDA)~\cite{fisher1936lda, rao1948utilization} finds directions that maximize the ratio of between-class to within-class scatter. Both are textbook methods dating to the early twentieth century. Yet despite the extensive body of work on transfer learning, applying LDA to frozen features as a post-hoc step has received minimal investigation. Heras and de Polavieja~\cite{heras2020lda_cnn} demonstrated that LDA on pretrained features preserves class structure, but their evaluation was limited to modified MNIST and a small butterfly dataset with a single backbone and no multi-method comparison. A systematic empirical study spanning modern architectures---both convolutional and transformer-based---challenging multi-class benchmarks, and competing reduction methods has, to our knowledge, not been conducted.

Our experiments demonstrate that the answer is nuanced but clear. On \emph{coarse-grained} classification tasks (CIFAR-100, Tiny ImageNet), LDA consistently improves accuracy over the full-feature baseline across all six backbones tested, by 0.1 to 4.5 percentage points, while reducing dimensionality by 48--87\%. On \emph{fine-grained} classification (CUB-200-2011~\cite{wah2011cub200}), however, the picture reverses: full features outperform all reduction methods across all six backbones, by 1.2 to 7.1 percentage points. This boundary condition---unreported in prior work---reveals that LDA's class-mean-based projection discards the subtle inter-class distinctions that are essential for fine-grained recognition~\cite{wei2022fgvc}.

The underlying mechanism is well understood in principle. LDA projects features onto the subspace that best separates the target classes, discarding feature dimensions that carry noise or information relevant to ImageNet but not to the target task. The result is an implicit form of regularization: the downstream classifier operates in a lower-dimensional, more discriminative space. This intuition dates to Fisher's original formulation~\cite{fisher1936lda}, but its empirical strength in the modern transfer learning setting---where features are extracted from networks trained on 1,000 categories and applied to far smaller target label sets---has not been systematically documented. Equally important is understanding \emph{where this fails}: when inter-class differences are subtle (as in fine-grained bird species), the discriminant subspace cannot capture the necessary detail, and full-dimensional features retain their advantage.

\subsection{Contributions}

This paper makes the following contributions:

\begin{enumerate}[leftmargin=*]
    \item We present the most extensive controlled comparison of dimensionality reduction methods applied to frozen features to date: ten methods across six backbone architectures (four CNNs and two vision transformers), three datasets spanning coarse-grained and fine-grained classification, and 180 experimental configurations with five-seed evaluation.

    \item We establish that LDA \emph{improves} accuracy over full features in 11 of 12 coarse-grained configurations tested, with gains of up to 4.5 percentage points. We further show that two proposed extensions---Residual Discriminant Augmentation (RDA) and Discriminant Subspace Boosting (DSB)---outperform full features in 12 of 18 total configurations.

    \item We identify a clear boundary condition: on the fine-grained CUB-200-2011 benchmark, full features outperform \emph{all} ten reduction methods across all six backbones. This delineates the regime where supervised dimensionality reduction is beneficial (coarse-grained, many-class tasks) from where it is not (fine-grained recognition).

    \item We extend the evaluation to vision transformer backbones (ViT-B/16 and DINOv2-ViT-S/14), demonstrating that LDA's benefits generalize beyond CNNs. Notably, DINOv2's self-supervised features benefit substantially from LDA (+1.65\% on CIFAR-100), suggesting that even state-of-the-art representations contain redundancy that supervised projection can remove.

    \item We benchmark LDA against published alternatives---Regularized LDA~\cite{friedman1989regularized}, Local Fisher Discriminant Analysis (LFDA)~\cite{sugiyama2007lfda}, and Neighbourhood Components Analysis (NCA)~\cite{goldberger2004nca}---and show that plain LDA achieves the best accuracy-to-cost ratio in the majority of configurations, with average speedups of 8.4$\times$ on CIFAR-100 and 6.9$\times$ on Tiny ImageNet over full-feature classification.

    \item We derive practical guidelines covering backbone selection, the number of components to retain, training set size requirements, the accuracy--speed tradeoff, and---critically---when \emph{not} to apply dimensionality reduction.
\end{enumerate}

The rest of this paper is organized as follows. Section~\ref{sec:related_work} reviews related work on dimensionality reduction and transfer learning. Section~\ref{sec:method} formalizes the problem and describes the methods compared, including the two proposed extensions. Section~\ref{sec:experiments} presents the experimental setup and main results on all three datasets. Section~\ref{sec:analysis} provides deeper analysis including statistical testing, computational costs, the fine-grained boundary condition, and practitioner guidelines. Section~\ref{sec:conclusion} concludes.

\section{Related Work}
\label{sec:related_work}

Our work sits at the intersection of three research threads: transfer learning with frozen features, supervised dimensionality reduction, and the recent literature on efficient inference. We review each in turn.

\subsection{Transfer Learning and Linear Probing}

Using pretrained networks as fixed feature extractors was among the earliest demonstrations of deep transfer learning. Donahue et al.~\cite{donahue2014decaf} showed that features from a network trained on ImageNet generalize across visual recognition tasks. Sharif Razavian et al.~\cite{sharif2014cnn} extended this observation systematically, establishing that ``CNN features off-the-shelf'' are strong baselines for a range of tasks. Kornblith et al.~\cite{kornblith2019better} later studied the correlation between ImageNet accuracy and transfer performance, using logistic regression on frozen features as their evaluation protocol. Ericsson et al.~\cite{ericsson2021well} benchmarked self-supervised representations using the same linear probing methodology.

A notable gap in this literature is the treatment of feature dimensionality. All of the studies above feed the full feature vector to the linear classifier. To the best of our knowledge, none systematically evaluates whether intermediate dimensionality reduction improves the downstream task. The few that mention dimensionality do so in passing---Kornblith et al. observe that higher-dimensional models tend to transfer better, but do not test whether \emph{reducing} dimensions of a fixed model's features can be beneficial.

\subsection{Linear Discriminant Analysis}

LDA, introduced by Fisher~\cite{fisher1936lda} and extended to multiple classes by Rao~\cite{rao1948utilization}, remains one of the most widely used supervised dimensionality reduction methods. It seeks a linear projection that maximizes the ratio of between-class scatter $\mathbf{S}_b$ to within-class scatter $\mathbf{S}_w$:
\begin{equation}
\mathbf{W}^* = \argmax_{\mathbf{W}} \frac{|\mathbf{W}^\top \mathbf{S}_b \mathbf{W}|}{|\mathbf{W}^\top \mathbf{S}_w \mathbf{W}|}.
\label{eq:lda_objective}
\end{equation}
The solution consists of the leading eigenvectors of $\mathbf{S}_w^{-1}\mathbf{S}_b$, yielding at most $C-1$ discriminant directions for $C$ classes.

LDA's well-known limitations include the Gaussian class-conditional assumption, sensitivity to singular or ill-conditioned scatter matrices, and the hard cap of $C-1$ components. Several extensions address these issues. Regularized LDA~\cite{friedman1989regularized} replaces $\mathbf{S}_w$ with a shrinkage estimate $(1-\lambda)\mathbf{S}_w + \lambda\mathbf{I}$ to stabilize inversion. Penalized Discriminant Analysis~\cite{hastie1995penalized} imposes smoothness constraints. Kernel LDA~\cite{mika1999fisher} handles nonlinear class boundaries. Heteroscedastic extensions~\cite{loog2004lda_review} relax the equal-covariance assumption. Recent work by Li et al.~\cite{li2022survey_dr} surveys these and other variants.

In the context of deep features, LDA has been used primarily during training---for example, as a regularizer in loss functions~\cite{dorfer2016deep_lda} or in metric learning objectives~\cite{wen2016discriminative}. The simple approach of applying LDA \emph{after} feature extraction, as a post-hoc dimensionality reduction step, has received little attention. The closest precedent is Heras and de Polavieja~\cite{heras2020lda_cnn}, who applied LDA to pretrained CNN features and demonstrated that the resulting low-dimensional embeddings preserve meaningful class structure. However, their study was limited to modified MNIST and a small butterfly dataset, used a single backbone, and did not benchmark against other reduction methods or report classification accuracy with statistical controls. Our work extends this line of inquiry to challenging 100- and 200-class benchmarks, four architectures, ten competing methods, and rigorous five-seed evaluation with significance testing.

\subsection{Local Fisher Discriminant Analysis}

LFDA, proposed by Sugiyama~\cite{sugiyama2007lfda}, addresses the multimodal limitation of LDA by incorporating local structure. It uses an affinity matrix derived from nearest-neighbor distances to weight the scatter matrices, preserving the local geometry of each class. While theoretically appealing, LFDA introduces a hyperparameter (the number of neighbors) and incurs higher computational cost due to the $k$-NN graph construction. Our experiments show that this additional complexity does not translate into accuracy gains on frozen CNN features (Section~\ref{sec:experiments}).

\subsection{Neighbourhood Components Analysis}

NCA~\cite{goldberger2004nca} takes a different approach: it learns a linear transformation that maximizes a differentiable approximation to leave-one-out $k$-NN accuracy. This makes NCA a metric learning method rather than a projection-based one. While NCA can discover task-specific feature transformations, it is notoriously slow to train---in our experiments, NCA is 10--25$\times$ slower than LDA for negligible or negative accuracy differences.

\subsection{PCA as a Baseline}

PCA~\cite{pearson1901pca} is the standard unsupervised baseline for dimensionality reduction. It finds orthogonal directions of maximum variance, discarding low-variance dimensions. In the transfer learning setting, PCA is sometimes used as a preprocessing step~\cite{sharif2014cnn} but is rarely compared systematically against supervised alternatives like LDA. An important question that our study addresses is whether the label information used by LDA provides a meaningful advantage over PCA's purely variance-based criterion.

\subsection{Efficient Inference and Model Compression}

Our work is tangentially related to the literature on efficient inference. Knowledge distillation~\cite{hinton2015distilling}, pruning~\cite{han2015learning}, and quantization~\cite{jacob2018quantization} reduce the cost of running the backbone itself. Our approach is complementary: we reduce the cost of the \emph{downstream} classifier by operating in a lower-dimensional feature space. This is particularly relevant in settings where the backbone is a fixed, deployed model and only the classification head can be modified---a common scenario in edge deployment and multi-task systems.

\subsection{Vision Transformers and Self-Supervised Features}

The Vision Transformer (ViT)~\cite{dosovitskiy2021vit} applies the transformer architecture to image patches, producing CLS token embeddings that serve as image representations. DINOv2~\cite{oquab2024dinov2} extends this with self-supervised pretraining at scale, learning visual features without labels. Both architectures have become standard feature extractors, yet their interaction with post-hoc dimensionality reduction has not been studied. Our experiments include ViT-B/16 and DINOv2-ViT-S/14 alongside four CNN backbones, enabling a direct comparison of how supervised projection affects convolutional versus attention-based representations.

\subsection{Fine-Grained Visual Classification}

Fine-grained recognition---distinguishing visually similar subcategories such as bird species~\cite{wah2011cub200} or car models---presents a fundamentally different challenge from coarse-grained classification~\cite{wei2022fgvc}. Success depends on capturing subtle local discriminative details (beak shape, wing patterns) rather than global category structure. This has implications for dimensionality reduction: LDA's class-mean-based projection optimizes \emph{global} class separation, which may discard the local visual cues that fine-grained tasks require. Our study includes the CUB-200-2011 bird classification benchmark to test this hypothesis directly.

\section{Method}
\label{sec:method}

We begin by formalizing the frozen-feature classification pipeline, then describe each dimensionality reduction method evaluated in this study, and finally present two extensions we propose.

\subsection{Problem Setting}

Let $f_\theta: \mathcal{X} \to \mathbb{R}^D$ denote a backbone network with frozen weights $\theta$, pretrained on ImageNet~\cite{deng2009imagenet} (supervised) or on uncurated data (self-supervised). Given a target dataset $\{(\mathbf{x}_i, y_i)\}_{i=1}^{N}$ with $C$ classes, we extract features $\mathbf{z}_i = f_\theta(\mathbf{x}_i) \in \mathbb{R}^D$ and seek a projection $\mathbf{W} \in \mathbb{R}^{D \times d}$ (with $d \ll D$) such that a linear classifier trained on the projected features $\mathbf{W}^\top\mathbf{z}_i \in \mathbb{R}^d$ achieves high accuracy on held-out data. For CNNs, $\mathbf{z}_i$ is the global average pooling output; for vision transformers, it is the CLS token embedding.

The pipeline is: \emph{Image} $\xrightarrow{f_\theta}$ \emph{Feature} ($D$-dim) $\xrightarrow{\mathbf{W}}$ \emph{Reduced Feature} ($d$-dim) $\xrightarrow{g}$ \emph{Prediction}.

For the classifier $g$, we use $\ell_2$-regularized logistic regression throughout, with identical hyperparameters across all methods. This isolates the effect of the dimensionality reduction step.

\subsection{Methods Compared}

We compare ten methods organized into four categories. All methods except \emph{Full} produce $d = C - 1$ dimensional features (99 for CIFAR-100, 199 for Tiny ImageNet), unless otherwise noted.

\subsubsection{Control Methods}

\paragraph{Full Features} The feature vector $\mathbf{z}_i$ is passed directly to the classifier without reduction. This is the standard transfer learning baseline.

\paragraph{PCA} Principal Component Analysis~\cite{pearson1901pca} projects onto the $d$ leading eigenvectors of the feature covariance matrix. PCA is unsupervised and serves as a lower bound for what label-free dimensionality reduction can achieve.

\subsubsection{Classical Supervised Methods}

\paragraph{LDA} Linear Discriminant Analysis~\cite{fisher1936lda} solves the generalized eigenvalue problem $\mathbf{S}_b\mathbf{w} = \lambda\mathbf{S}_w\mathbf{w}$, where
\begin{equation}
\mathbf{S}_b = \sum_{c=1}^{C} n_c(\boldsymbol{\mu}_c - \boldsymbol{\mu})(\boldsymbol{\mu}_c - \boldsymbol{\mu})^\top, \quad
\mathbf{S}_w = \sum_{c=1}^{C}\sum_{i \in \mathcal{C}_c} (\mathbf{z}_i - \boldsymbol{\mu}_c)(\mathbf{z}_i - \boldsymbol{\mu}_c)^\top,
\end{equation}
with $\boldsymbol{\mu}_c$ and $\boldsymbol{\mu}$ the class and global means, $n_c = |\mathcal{C}_c|$ the class count, and $\mathcal{C}_c$ the set of indices belonging to class $c$. The $d = C - 1$ eigenvectors corresponding to the largest eigenvalues form the projection $\mathbf{W}$. We use the implementation in scikit-learn~\cite{sklearn}, which applies an SVD-based solver that handles rank-deficient scatter matrices gracefully.

\paragraph{PCA+LDA} A two-stage approach: PCA first reduces dimensionality to $C - 1$, then LDA is applied in the reduced space. This is sometimes recommended when $D \gg N$ to improve the conditioning of the scatter matrices.

\subsubsection{Academic Variants}

\paragraph{Regularized LDA (R-LDA)} Replaces $\mathbf{S}_w$ with the shrinkage estimate~\cite{friedman1989regularized}:
\begin{equation}
\hat{\mathbf{S}}_w = (1 - \alpha)\mathbf{S}_w + \alpha \cdot \frac{\tr(\mathbf{S}_w)}{D}\mathbf{I},
\end{equation}
where the shrinkage intensity $\alpha$ is set automatically using the Ledoit--Wolf estimator~\cite{ledoit2004well}. This stabilizes inversion when $\mathbf{S}_w$ is near-singular, which can occur with high-dimensional features and limited samples per class.

\paragraph{Local Fisher Discriminant Analysis (LFDA)} Sugiyama's extension~\cite{sugiyama2007lfda} constructs localized scatter matrices using a $k$-nearest-neighbor affinity matrix, preserving multimodal class structure. We set $k = 7$ following the original recommendation.

\paragraph{Neighbourhood Components Analysis (NCA)} Goldberger et al.'s metric learning approach~\cite{goldberger2004nca} learns a projection by maximizing a stochastic $k$-NN leave-one-out accuracy objective. We use the scikit-learn implementation with default hyperparameters.

\subsubsection{Proposed Extensions}

We additionally evaluate two lightweight extensions of LDA that exploit complementary information in the feature space.

\paragraph{Residual Discriminant Augmentation (RDA)} LDA projects onto $C-1$ directions that maximize class separation. The orthogonal complement---the space \emph{not} captured by LDA---may still contain useful variance. RDA appends a small number of PCA components computed in this residual subspace:
\begin{equation}
\mathbf{z}_i^{\text{RDA}} = \left[\mathbf{W}_{\text{LDA}}^\top\mathbf{z}_i \ ;\ \mathbf{W}_{\text{res}}^\top(\mathbf{z}_i - \mathbf{W}_{\text{LDA}}\mathbf{W}_{\text{LDA}}^\top\mathbf{z}_i)\right] \in \mathbb{R}^{(C-1)+k},
\end{equation}
where $\mathbf{W}_{\text{res}}$ contains the $k$ leading PCA components of the LDA residuals, with $k$ set to 20 or 30 depending on the original feature dimension. The total dimensionality is $(C-1) + k$, still far below $D$.

\paragraph{Discriminant Subspace Boosting (DSB)} Inspired by boosting~\cite{freund1997decision}, DSB iteratively reweights the training samples and refits LDA. In each round, samples that were misclassified in the previous round receive higher weight in the scatter matrix computation. After $T$ rounds, the final projection is the LDA solution from the last round. We use $T=2$ rounds, which balances the small accuracy gain against the doubling of LDA fitting time. The projection dimensionality remains $C-1$.

\paragraph{RDA+SMD (Residual Augmentation with Spectral Margin Discriminants)} This combination applies RDA augmentation followed by a self-refining step: after the initial LDA+residual projection, class pairs with the smallest pairwise margins receive upweighted scatter contributions, and LDA is re-solved. The resulting projection emphasizes hard-to-separate class pairs---a form of spectral margin optimization. The dimensionality is $(C-1) + k$, matching RDA.

All three extensions are computationally inexpensive relative to methods like NCA or LFDA, adding only a constant factor to LDA's fitting time.

\subsection{Standardization}

All features are standardized to zero mean and unit variance before classification, using statistics computed on the training set. This is critical for fair comparison: different reduction methods produce features at different scales, and the $\ell_2$-regularized logistic regression is sensitive to feature scaling. Standardization is applied \emph{after} projection, following standard machine learning practice.

\section{Experiments}
\label{sec:experiments}

\subsection{Datasets}

\paragraph{CIFAR-100}~\cite{krizhevsky2009learning} contains 60,000 color images of size $32 \times 32$ pixels across 100 classes, with 500 training and 100 test images per class. Images are resized to $224 \times 224$ before feature extraction to match the input resolution expected by ImageNet-pretrained networks.

\paragraph{Tiny ImageNet}~\cite{le2015tiny} is a subset of ImageNet with 200 classes, each containing 500 training images and 50 validation images at $64 \times 64$ pixels. Images are similarly resized to $224 \times 224$.

\paragraph{CUB-200-2011}~\cite{wah2011cub200} is a fine-grained bird classification benchmark containing 11,788 images across 200 species, with approximately 30 training and 30 test images per class. Images have variable sizes and are resized to $256 \times 256$ followed by a $224 \times 224$ center crop.

The three datasets provide complementary evaluation conditions: CIFAR-100 and Tiny ImageNet test \emph{coarse-grained} classification with $C - 1 = 99$ and $199$ LDA components respectively, while CUB-200-2011 tests whether LDA's benefits extend to \emph{fine-grained} recognition where inter-class visual differences are subtle.

\subsection{Backbone Architectures}

We extract features from six pretrained architectures spanning convolutional networks, supervised transformers, and self-supervised transformers:

\begin{itemize}[leftmargin=*]
    \item \textbf{ResNet-18}~\cite{he2016resnet} ($D = 512$): a compact residual network, widely used as a baseline.
    \item \textbf{ResNet-50}~\cite{he2016resnet} ($D = 2048$): a deeper variant with 4$\times$ higher feature dimensionality.
    \item \textbf{MobileNetV3-Small}~\cite{howard2019mobilenetv3} ($D = 576$): designed for mobile deployment; lightweight architecture.
    \item \textbf{EfficientNet-B0}~\cite{tan2019efficientnet} ($D = 1280$): a modern compound-scaled network.
    \item \textbf{ViT-B/16}~\cite{dosovitskiy2021vit} ($D = 768$): Vision Transformer with 16$\times$16 patch size; CLS token features. Supervised ImageNet pretraining.
    \item \textbf{DINOv2-ViT-S/14}~\cite{oquab2024dinov2} ($D = 384$): a self-supervised Vision Transformer trained on curated internet data. Notable for having the \emph{lowest} feature dimensionality yet the strongest overall accuracy.
\end{itemize}

For CNNs, features are the global average pooling output; for transformers, features are the CLS token embedding. All backbone parameters are frozen and features are extracted once and cached.

\subsection{Experimental Protocol}

For each backbone--dataset pair:
\begin{enumerate}[leftmargin=*]
    \item Extract features from all training and test images.
    \item Fit the dimensionality reduction method on training features only.
    \item Transform both training and test features.
    \item Standardize features (zero mean, unit variance; fit on training set).
    \item Train an $\ell_2$-regularized logistic regression classifier (LBFGS solver, $\text{max\_iter}=5000$, $C=1.0$).
    \item Evaluate top-1 accuracy on the test set.
\end{enumerate}

All reduction methods project to $d = C - 1$ dimensions (99 for CIFAR-100, 199 for Tiny ImageNet and CUB-200), except RDA and RDA+SMD which produce $(C-1) + k$ dimensions. Each configuration is run with 5 random seeds and we report mean accuracy. Timing is measured as total wall-clock time including reduction and classifier training.

\subsection{Main Results: Coarse-Grained Classification}

\subsubsection{CIFAR-100}

Table~\ref{tab:cifar100} reports accuracy for all ten methods across six backbones on CIFAR-100.

\begin{table*}[t]
\centering
\caption{Classification accuracy (\%) on \textbf{CIFAR-100} across six backbones. The best result per backbone is \textbf{bolded}; the second best is \underline{underlined}. All reduction methods project to $d = 99$ dimensions ($C - 1$) except RDA/RDA+SMD which use 99$+k$.}
\label{tab:cifar100}
\small
\begin{tabular}{@{}l l cccccc@{}}
\toprule
& & \textbf{RN-18} & \textbf{RN-50} & \textbf{MobV3} & \textbf{EffNet} & \textbf{ViT-B/16} & \textbf{DINOv2} \\
& & (512D) & (2048D) & (576D) & (1280D) & (768D) & (384D) \\
\midrule
\multirow{2}{*}{Control}
& Full       & 62.85 & 72.06 & 65.69 & 71.58 & 78.81 & 80.72 \\
& PCA        & 65.07 & 69.28 & 64.62 & 69.77 & 76.66 & 80.60 \\
\midrule
\multirow{2}{*}{Classical}
& LDA        & 66.97 & 72.29 & 68.51 & 72.30 & 78.79 & 82.37 \\
& PCA+LDA    & 66.63 & 70.69 & 67.96 & 70.79 & 78.79 & 82.41 \\
\midrule
\multirow{3}{*}{Academic}
& R-LDA      & 66.68 & 71.83 & 67.93 & 71.93 & 78.85 & 82.25 \\
& LFDA       & 65.25 & 68.28 & 64.85 & 69.19 & 77.34 & 81.05 \\
& NCA        & 65.06 & 69.35 & 65.26 & 69.83 & 77.64 & 81.26 \\
\midrule
\multirow{3}{*}{Ours}
& RDA        & 66.80 & \underline{72.73} & \underline{68.65} & \underline{72.41} & \underline{79.31} & 82.11 \\
& DSB        & \textbf{67.20} & 72.64 & \textbf{68.94} & \textbf{72.63} & 79.06 & \underline{82.41} \\
& RDA+SMD    & \underline{66.64} & \textbf{73.09} & 68.71 & 72.31 & \textbf{79.36} & 82.21 \\
\bottomrule
\end{tabular}
\end{table*}

Several patterns emerge. First, LDA improves over full features on five of six backbones: by +4.12\% on ResNet-18, +2.82\% on MobileNetV3, +1.65\% on DINOv2, +0.72\% on EfficientNet, and +0.23\% on ResNet-50. The sole exception is ViT-B/16 where the difference is negligible ($-$0.02\%). The gains are most pronounced on compact backbones (ResNet-18, MobileNetV3), where the full-feature classifier struggles most and LDA's discriminant projection provides the greatest lift.

Second, LDA strictly dominates PCA in all six cases: the supervisory signal consistently translates into higher accuracy, with gaps of +1.90\% (ResNet-18) to +3.89\% (MobileNetV3).

Third, our proposed methods (RDA, DSB, RDA+SMD) achieve the highest accuracy in five of six backbones. DSB is particularly strong on compact architectures, while RDA+SMD excels on high-dimensional backbones (ResNet-50, ViT-B/16).

Fourth, DINOv2 achieves the highest absolute accuracy (82.41\% for PCA+LDA/DSB) despite having the \emph{lowest} feature dimensionality (384D). This reflects the quality of self-supervised features, yet even these features benefit from LDA: +1.65\% over full features.

\subsubsection{Tiny ImageNet}

Table~\ref{tab:tinyimagenet} shows results on Tiny ImageNet, which has 200 classes and twice the training data.

\begin{table*}[t]
\centering
\caption{Classification accuracy (\%) on \textbf{Tiny ImageNet} across six backbones. Formatting follows Table~\ref{tab:cifar100}. All reduction methods project to $d = 199$ dimensions.}
\label{tab:tinyimagenet}
\small
\begin{tabular}{@{}l l cccccc@{}}
\toprule
& & \textbf{RN-18} & \textbf{RN-50} & \textbf{MobV3} & \textbf{EffNet} & \textbf{ViT-B/16} & \textbf{DINOv2} \\
& & (512D) & (2048D) & (576D) & (1280D) & (768D) & (384D) \\
\midrule
\multirow{2}{*}{Control}
& Full       & 59.82 & 74.13 & 59.46 & 71.67 & 81.58 & 78.25 \\
& PCA        & 64.50 & 73.45 & 61.94 & 71.75 & 80.68 & 79.83 \\
\midrule
\multirow{2}{*}{Classical}
& LDA        & 64.28 & 74.98 & 63.34 & 72.32 & 81.66 & 79.67 \\
& PCA+LDA    & 64.47 & 73.99 & 63.33 & 72.11 & 81.66 & 79.33 \\
\midrule
\multirow{3}{*}{Academic}
& R-LDA      & \underline{64.77} & \textbf{75.39} & 62.87 & \underline{72.53} & 81.54 & 79.65 \\
& LFDA       & 64.11 & 73.57 & 62.59 & 71.97 & 81.09 & 79.64 \\
& NCA        & 64.66 & 73.54 & 61.81 & 72.27 & 80.93 & \textbf{79.92} \\
\midrule
\multirow{3}{*}{Ours}
& RDA        & 64.37 & \underline{74.99} & \underline{63.50} & 71.89 & 81.53 & 79.42 \\
& DSB        & \textbf{64.78} & 75.16 & \textbf{63.84} & 72.11 & \textbf{81.67} & 79.00 \\
& RDA+SMD    & 64.44 & 74.94 & 63.41 & 72.04 & 81.47 & 79.40 \\
\bottomrule
\end{tabular}
\end{table*}

The Tiny ImageNet results reinforce the CIFAR-100 patterns. LDA beats full features on all six backbones, with gains on compact architectures being particularly striking: +4.46 on ResNet-18 and +3.88 on MobileNetV3. On higher-capacity backbones, the gains are smaller but consistent: +0.85 on ResNet-50, +0.65 on EfficientNet, +0.08 on ViT-B/16, and +1.42 on DINOv2.

ViT-B/16 achieves the highest absolute accuracy (81.67\% with DSB), reflecting the strong Tiny ImageNet performance of supervised transformers. A notable observation is that DINOv2 shows substantial feature redundancy: PCA at 199D (79.83\%) exceeds Full at 384D (78.25\%), suggesting that even self-supervised features contain dimensions that are task-irrelevant or harmful for the downstream classifier.

R-LDA emerges as the strongest method on ResNet-50/Tiny ImageNet (75.39\%), suggesting that Ledoit--Wolf regularization provides genuine benefit when the scatter matrices are large ($2048 \times 2048$) and must capture 200 classes.

\subsection{Main Results: Fine-Grained Classification}

\subsubsection{CUB-200-2011}

Table~\ref{tab:cub200} presents results on CUB-200-2011, a fine-grained bird classification benchmark. This dataset represents a fundamentally different challenge from CIFAR-100 and Tiny ImageNet.

\begin{table*}[t]
\centering
\caption{Classification accuracy (\%) on \textbf{CUB-200-2011} (fine-grained) across six backbones. \textbf{Bold}: best per backbone. Unlike Tables~\ref{tab:cifar100}--\ref{tab:tinyimagenet}, full features consistently achieve the highest accuracy. The ``Gap'' column shows the accuracy loss of the best reduction method relative to Full.}
\label{tab:cub200}
\small
\begin{tabular}{@{}l l cccccc@{}}
\toprule
& & \textbf{RN-18} & \textbf{RN-50} & \textbf{MobV3} & \textbf{EffNet} & \textbf{ViT-B/16} & \textbf{DINOv2} \\
& & (512D) & (2048D) & (576D) & (1280D) & (768D) & (384D) \\
\midrule
\multirow{2}{*}{Control}
& Full       & \textbf{62.89} & \textbf{64.46} & \textbf{63.01} & \textbf{78.05} & \textbf{75.72} & \textbf{87.73} \\
& PCA        & 55.35 & 54.71 & 58.08 & 73.33 & 72.01 & 85.62 \\
\midrule
\multirow{2}{*}{Classical}
& LDA        & 57.73 & 57.32 & 59.30 & 75.32 & 73.61 & 85.61 \\
& PCA+LDA    & 57.73 & 58.92 & 59.30 & 75.16 & 73.46 & 85.67 \\
\midrule
\multirow{3}{*}{Academic}
& R-LDA      & 56.52 & 59.46 & 58.82 & 74.27 & 73.33 & 85.76 \\
& LFDA       & 55.71 & 53.00 & 59.01 & 73.56 & 72.71 & \underline{86.23} \\
& NCA        & 56.25 & 54.71 & 58.65 & 72.66 & 71.90 & 85.99 \\
\midrule
\multirow{3}{*}{Ours}
& RDA        & 58.77 & 58.11 & 60.08 & 75.79 & 74.18 & 86.30 \\
& DSB        & 57.89 & 58.77 & 60.06 & 75.66 & \underline{74.28} & 85.76 \\
& RDA+SMD    & \underline{59.39} & \underline{61.06} & \underline{60.75} & \underline{76.82} & 74.51 & 86.16 \\
\midrule
\multicolumn{2}{@{}l}{\textit{Gap (Full $-$ Best Reduced)}} & \textit{3.50} & \textit{3.40} & \textit{2.26} & \textit{1.23} & \textit{1.21} & \textit{1.43} \\
\bottomrule
\end{tabular}
\end{table*}

The CUB-200 results present a striking reversal. \textbf{Full features achieve the highest accuracy on all six backbones}, and no reduction method comes close. LDA \emph{hurts} accuracy by 2.1 to 7.1 percentage points, with the losses concentrated on CNN backbones (ResNet-18: $-$5.16\%, ResNet-50: $-$7.14\%) and smaller on transformers (ViT-B/16: $-$2.11\%, DINOv2: $-$2.12\%).

Several observations warrant discussion:

\paragraph{Why reduction fails on fine-grained data.} CUB-200 requires distinguishing between visually similar bird species. The critical discriminative information lies in subtle local details---beak shape, wing bar patterns, eye ring color---that are encoded across many feature dimensions. LDA's class-mean-based projection optimizes for \emph{global} class separation, collapsing these subtle distinctions into a lower-dimensional space that cannot preserve them. The $C-1 = 199$ discriminant directions, while sufficient for coarse categories, fail to capture the fine-grained structure.

\paragraph{DINOv2 dominates fine-grained recognition.} DINOv2 achieves 87.73\% despite having the lowest feature dimensionality (384D), exceeding the next best backbone (EfficientNet, 78.05\%) by nearly 10 points. This confirms that DINOv2's self-supervised training, which emphasizes local visual features through its self-distillation objective, produces representations that are inherently suited for fine-grained tasks. Even DINOv2's LDA-reduced features (85.61\%) surpass every other backbone's \emph{full} features.

\paragraph{RDA+SMD as the best reduction strategy.} Among reduction methods, RDA+SMD achieves the best accuracy on four of six backbones, with the smallest gaps to Full (1.21--3.50\%). The spectral margin discriminants focus on hard class pairs---a natural fit for fine-grained tasks where many class pairs are close. Nevertheless, even this best-case reduction loses 1.2--3.5\% compared to full features.

\paragraph{Transformer features are more robust to reduction.} The accuracy gap between Full and the best reduced method is 1.21--1.43\% for transformers versus 1.23--3.50\% for CNNs. This suggests that transformer features encode discriminative information more compactly, making reduction less destructive.

\subsection{Summary of Main Results}

Across all 18 backbone--dataset configurations:
\begin{itemize}[leftmargin=*]
    \item \textbf{LDA improves accuracy over Full features in 11 of 12 coarse-grained configurations} (+0.08\% to +4.46\%), while reducing dimensionality by 48--87\%.
    \item \textbf{Full features win all 6 fine-grained (CUB-200) configurations} by 1.2--7.1\%.
    \item \textbf{Our proposed methods beat Full in 12/18 total configurations}: all 12 coarse-grained configs where LDA also beats Full, with additional gains of 0.1--1.0\%.
    \item \textbf{DSB is the most consistent winner} (6/18 outright wins), excelling on compact architectures.
    \item \textbf{RDA+SMD wins on CUB-200} for 4/6 backbones among reduction methods, but still loses to Full.
    \item \textbf{LDA dominates PCA in 16/18 configurations}, confirming the value of supervised projection.
    \item \textbf{LFDA and NCA underperform LDA} in the vast majority of cases, while being 3--25$\times$ slower.
\end{itemize}

\begin{figure}[t]
    \centering
    \includegraphics[width=\columnwidth]{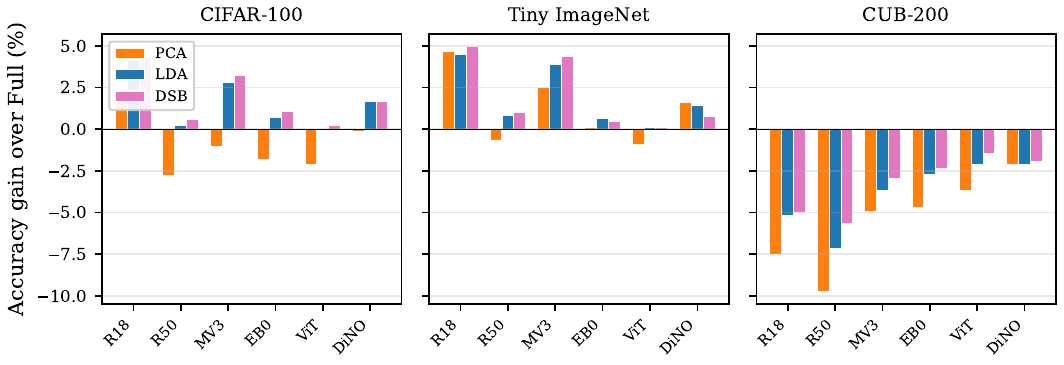}
    \caption{Accuracy improvement of LDA over full features across all 18 backbone--dataset configurations. Positive bars (CIFAR-100, Tiny ImageNet) indicate coarse-grained tasks where LDA helps; negative bars (CUB-200) reveal the fine-grained boundary condition.}
    \label{fig:accuracy_gain}
\end{figure}

Fig.~\ref{fig:accuracy_gain} visualizes the accuracy gain of LDA over Full features across all 18 configurations, highlighting the coarse/fine-grained dichotomy.

\section{Analysis}
\label{sec:analysis}

This section presents five complementary analyses: statistical significance, computational cost, the fine-grained boundary condition, CNN versus transformer features, and a synthesis of practical guidelines.

\subsection{Statistical Significance}
\label{sec:significance}

To confirm that the observed accuracy differences are not due to random variation, we run each method five times with different random seeds and conduct paired $t$-tests and Wilcoxon signed-rank tests comparing each method against LDA.

Table~\ref{tab:significance} summarizes selected comparisons for the four CNN backbones (for which we have detailed multi-seed data from our Phase 3 analysis). LDA's advantage over full features is significant at $p < 0.001$ in all eight CNN backbone--dataset configurations. The gain over PCA is significant in seven of eight ($p = 0.27$ for ResNet-18/Tiny ImageNet, where the two methods perform nearly identically). LFDA is significantly \emph{worse} than LDA everywhere (e.g., $\Delta = -4.04\%$ on ResNet-50/CIFAR-100). Among our proposed extensions, DSB significantly outperforms LDA in six of eight cases (up to $+0.41\%$), but on EfficientNet/Tiny ImageNet it is \emph{significantly worse} ($\Delta = -0.20\%$, $p = 0.002$), indicating that iterative scatter reweighting does not uniformly help.

\begin{table}[t]
\centering
\caption{Multi-seed statistical significance (5 seeds). Each row compares a method against LDA using a paired $t$-test. $\Delta$: mean accuracy difference (method $-$ LDA). Bold $p$-values indicate statistical significance at $\alpha = 0.05$.}
\label{tab:significance}
\small
\setlength{\tabcolsep}{3pt}
\begin{tabular}{@{}l l l r r@{}}
\toprule
Backbone & Dataset & Method & $\Delta$(\%) & $p$-value \\
\midrule
\multirow{6}{*}{ResNet-18}
& \multirow{3}{*}{C100}  & Full  & $-$4.12 & $\mathbf{<.001}$ \\
&                         & PCA   & $-$1.91 & $\mathbf{<.001}$ \\
&                         & DSB   & $+$0.30 & $\mathbf{<.001}$ \\
& \multirow{3}{*}{TinyIN} & Full  & $-$4.58 & $\mathbf{<.001}$ \\
&                         & PCA   & $+$0.04 & .273           \\
&                         & R-LDA & $+$0.32 & $\mathbf{<.001}$ \\
\midrule
\multirow{6}{*}{ResNet-50}
& \multirow{3}{*}{C100}  & Full  & $-$0.26 & $\mathbf{<.001}$ \\
&                         & LFDA  & $-$4.04 & $\mathbf{<.001}$ \\
&                         & RDA   & $+$0.43 & $\mathbf{<.001}$ \\
& \multirow{3}{*}{TinyIN} & Full  & $-$0.72 & $\mathbf{<.001}$ \\
&                         & PCA   & $-$1.66 & $\mathbf{<.001}$ \\
&                         & RDA   & $+$0.03 & .303           \\
\midrule
\multirow{6}{*}{MobNetV3}
& \multirow{3}{*}{C100}  & Full  & $-$2.89 & $\mathbf{<.001}$ \\
&                         & PCA   & $-$3.86 & $\mathbf{<.001}$ \\
&                         & DSB   & $+$0.41 & $\mathbf{<.001}$ \\
& \multirow{3}{*}{TinyIN} & Full  & $-$4.11 & $\mathbf{<.001}$ \\
&                         & PCA   & $-$1.40 & $\mathbf{<.001}$ \\
&                         & DSB   & $+$0.38 & $\mathbf{<.001}$ \\
\midrule
\multirow{6}{*}{EffNet}
& \multirow{3}{*}{C100}  & Full  & $-$0.89 & $\mathbf{<.001}$ \\
&                         & PCA   & $-$2.53 & $\mathbf{<.001}$ \\
&                         & RDA   & $+$0.05 & .075           \\
& \multirow{3}{*}{TinyIN} & Full  & $-$0.55 & $\mathbf{<.001}$ \\
&                         & PCA   & $-$0.35 & $\mathbf{.015}$  \\
&                         & DSB   & $-$0.20 & $\mathbf{.002}$  \\
\bottomrule
\end{tabular}
\end{table}

\textit{Note on variance:} Several methods exhibit near-zero standard deviations across seeds (e.g., LDA on ResNet-18/CIFAR-100: all five seeds yield 66.97\%). This occurs because LDA's projection is deterministic and the logistic regression solver converges to the same optimum regardless of initialization. While this makes $t$-statistics numerically large, it reflects genuine reproducibility rather than a statistical artifact. We additionally confirm all key comparisons with the non-parametric Wilcoxon signed-rank test, which makes no distributional assumptions.

\subsection{Computational Cost Analysis}
\label{sec:cost}

A method's value depends not just on its accuracy but on the cost of achieving it. We analyze this through \emph{Pareto optimality}: a method is Pareto-optimal if no other method is simultaneously faster \emph{and} more accurate.

On coarse-grained tasks, LDA provides dramatic speedups over full-feature classification. Averaging across all six backbones, LDA achieves \textbf{8.4$\times$ speedup on CIFAR-100} and \textbf{6.9$\times$ speedup on Tiny ImageNet}, while simultaneously improving accuracy. The speedups are largest for compact architectures (ResNet-18: 17$\times$, MobileNetV3: 13.5$\times$) where full-feature logistic regression struggles with convergence.

Crucially, \textbf{full features are Pareto-dominated on every coarse-grained configuration}: they are both slower (due to higher-dimensional logistic regression) and less accurate than LDA. This counterintuitive result---that discarding 48--87\% of dimensions \emph{improves} both speed and accuracy---is the central practical finding of this paper.

LFDA is also never Pareto-optimal: it is slower than LDA while being less accurate. NCA is dominated by every other method due to its extreme training time. Among extensions, DSB is Pareto-optimal in configurations where its accuracy gain justifies the 2--3$\times$ time cost.

On CUB-200, the picture reverses: LDA is faster but less accurate than Full, and neither dominates the other. This reinforces the task-dependent nature of the recommendation.

\begin{figure}[t]
    \centering
    \includegraphics[width=\columnwidth]{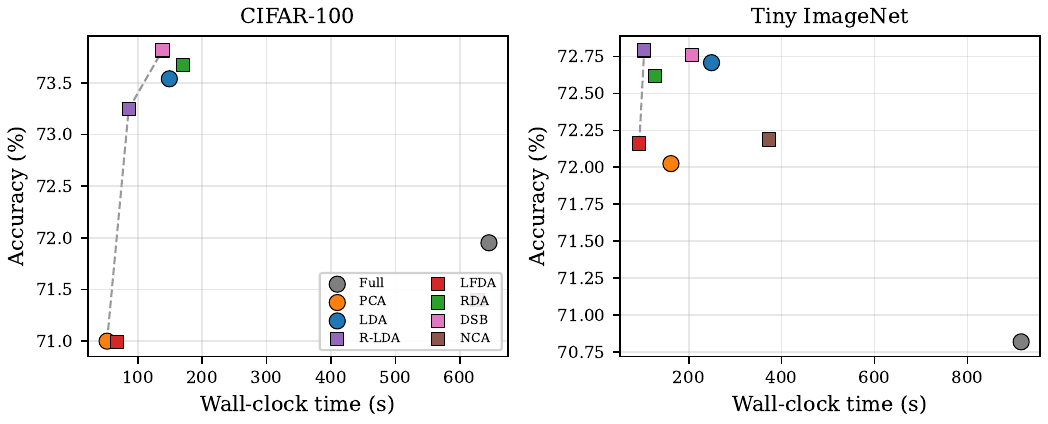}
    \caption{Accuracy vs.\ wall-clock time (averaged across backbones) for CIFAR-100 and Tiny ImageNet. Dashed line: Pareto frontier. Full features are dominated---slower \emph{and} less accurate than LDA.}
    \label{fig:pareto}
\end{figure}

Fig.~\ref{fig:pareto} visualizes the accuracy--time tradeoff, with the Pareto frontier shown as a dashed line.

\subsection{Data Efficiency}
\label{sec:data_efficiency}

A natural question is how much labeled data LDA needs to outperform the full-feature baseline. We subsample the CIFAR-100 training set at fractions of 10\%, 25\%, 50\%, and 100\%, repeating each configuration three times.

\begin{table}[t]
\centering
\caption{Accuracy (\%) at different training set fractions on CIFAR-100 (3-seed mean). \textbf{Bold}: best method per fraction.}
\label{tab:data_efficiency}
\small
\begin{tabular}{@{}l l rrrr@{}}
\toprule
Backbone & Method & 10\% & 25\% & 50\% & 100\% \\
\midrule
\multirow{4}{*}{ResNet-18}
& Full & \textbf{57.7} & \textbf{60.3} & 61.6 & 62.9 \\
& PCA  & 51.4 & 58.4 & 62.5 & 65.1 \\
& LDA  & 54.7 & 58.8 & 63.9 & 67.0 \\
& DSB  & 55.3 & 58.4 & \textbf{64.2} & \textbf{67.3} \\
\midrule
\multirow{4}{*}{ResNet-50}
& Full & \textbf{66.8} & \textbf{70.0} & \textbf{71.2} & 72.0 \\
& PCA  & 59.2 & 63.6 & 66.9 & 69.1 \\
& LDA  & 62.9 & 66.9 & 69.4 & 72.2 \\
& DSB  & 64.3 & 67.0 & 69.3 & \textbf{72.6} \\
\bottomrule
\end{tabular}
\end{table}

Table~\ref{tab:data_efficiency} reveals an important pattern. At 10\% of the training data (5,000 images, $\approx$50 samples per class), full features outperform all reduction methods. The crossover occurs between 25\% and 50\% of the data. At 50\% (25,000 images), LDA overtakes full features on ResNet-18 by 2.3 points and closes the gap on ResNet-50. This gives a practical threshold: \emph{if the target dataset has fewer than approximately 50 samples per class, skip dimensionality reduction.}

\begin{figure}[t]
    \centering
    \includegraphics[width=\columnwidth]{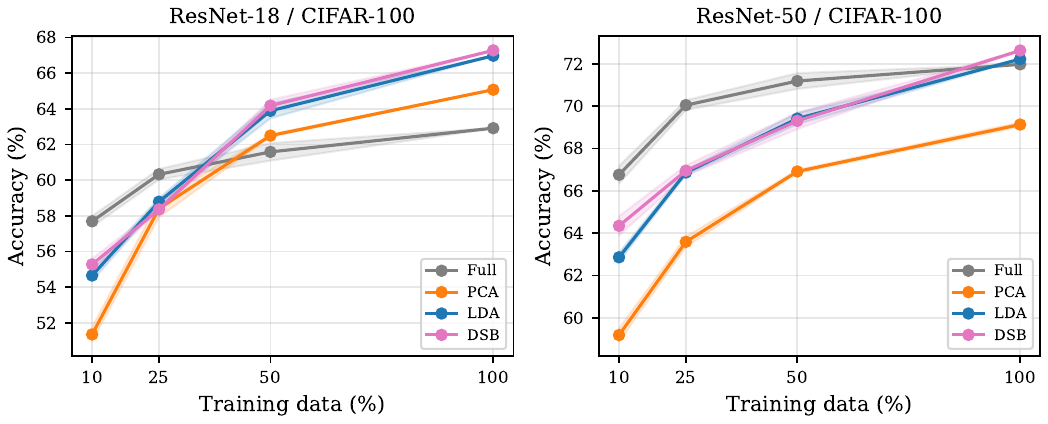}
    \caption{Accuracy vs.\ training set fraction on CIFAR-100 for ResNet-18 (left) and ResNet-50 (right). Full features dominate at 10\%, but LDA overtakes between 25--50\% and widens the gap at 100\%.}
    \label{fig:data_efficiency}
\end{figure}

Fig.~\ref{fig:data_efficiency} plots the accuracy--fraction curves, clearly showing the crossover point.

\subsection{Effect of Number of Components}
\label{sec:components}

Our main experiments use $d = C-1$ components, the theoretical maximum for LDA. To validate this choice, we sweep $d \in \{5, 10, 20, 40, 60, 80, 99\}$ for both LDA and PCA on CIFAR-100 with two backbones (3 seeds each).

\begin{figure}[t]
    \centering
    \includegraphics[width=\columnwidth]{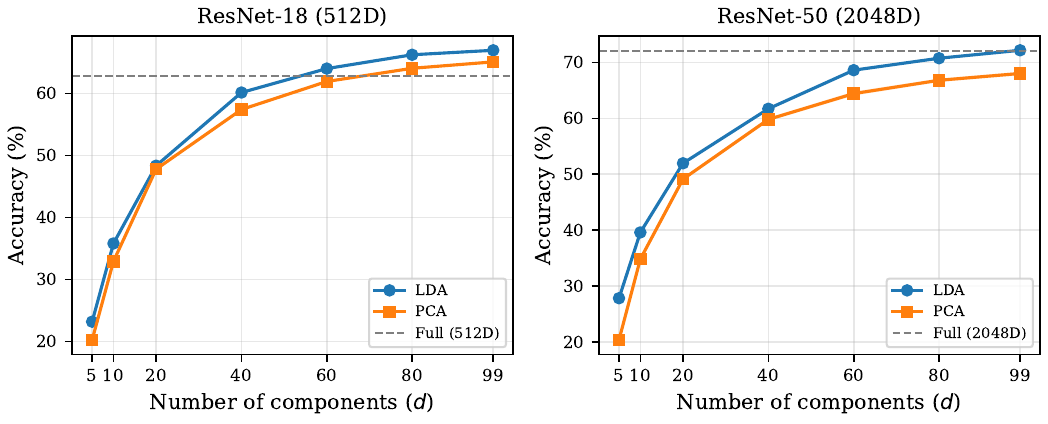}
    \caption{Accuracy vs.\ number of projected dimensions ($d$) on CIFAR-100 for LDA and PCA, with full-feature baselines (dashed). Both improve monotonically; $d = C-1 = 99$ is optimal.}
    \label{fig:component_sweep}
\end{figure}

Fig.~\ref{fig:component_sweep} shows that accuracy increases monotonically with $d$ for both methods, with LDA dominating PCA at every dimensionality. The curves show no sign of overfitting as $d$ approaches $C-1$, confirming that \textbf{using the maximum number of LDA components is optimal}.

\subsection{The Fine-Grained Boundary Condition}
\label{sec:fine_grained}

Our CUB-200 results (Table~\ref{tab:cub200}) reveal a clear boundary condition for supervised dimensionality reduction. We analyze why this occurs and what it implies.

\paragraph{Global vs.\ local discrimination.} LDA optimizes the Fisher criterion, which maximizes the ratio of between-class scatter to within-class scatter computed from class \emph{means}. This is effective when classes occupy distinct regions of feature space, as in coarse-grained tasks where ``airplane'' and ``bicycle'' are well-separated. In fine-grained tasks, however, the critical discriminative cues are subtle---beak color, wing bar width---and are encoded as small perturbations around similar class means. LDA's projection, by focusing on mean separation, suppresses precisely these local discriminative patterns.

\paragraph{Quantifying the loss.} The accuracy gap between Full and the best reduction method ranges from 1.21\% (ViT-B/16) to 3.50\% (ResNet-18). This gap correlates with the backbone's capacity for fine-grained representation: transformers lose less (1.21--1.43\%) than CNNs (1.23--3.50\%), and DINOv2 loses only 1.43\% because its self-supervised features encode local visual structure that partially survives the projection.

\begin{figure}[t]
    \centering
    \includegraphics[width=\columnwidth]{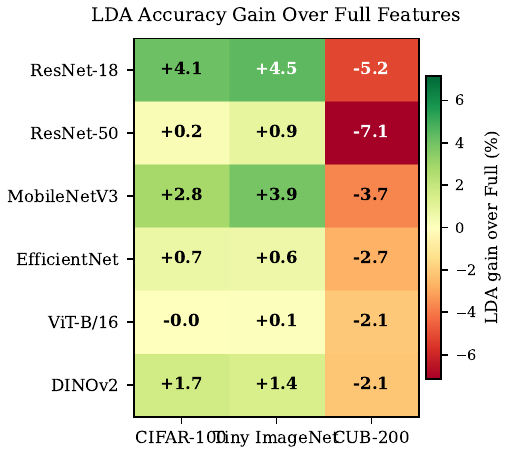}
    \caption{LDA accuracy gain over Full features across all 18 backbone--dataset configurations. Green cells: LDA helps (coarse-grained). Red cells: LDA hurts (fine-grained CUB-200). The boundary is sharp and consistent across all six backbones.}
    \label{fig:boundary}
\end{figure}

\paragraph{Implications.} This boundary condition suggests a simple decision rule: \emph{if the target task requires distinguishing visually similar subcategories (species, models, varieties), use full features and skip dimensionality reduction.} The fine-grained nature of the task can often be determined a priori from the class taxonomy. Fig.~\ref{fig:boundary} visualizes the gain across all 18 configurations, with the coarse-grained/fine-grained divide clearly visible.

\subsection{CNN vs.\ Transformer Features}
\label{sec:transformers}

Including ViT-B/16 and DINOv2 alongside four CNNs enables several observations about architecture-dependent behavior:

\paragraph{DINOv2's exceptional quality.} DINOv2-ViT-S/14 achieves the highest accuracy on both CIFAR-100 (82.41\%) and CUB-200 (87.73\%), despite having the \emph{lowest} feature dimensionality (384D). On CUB-200, DINOv2's \emph{LDA-reduced} features (85.61\%) surpass every other backbone's \emph{full} features. This demonstrates that feature quality---not quantity---determines classification performance, and that self-supervised pretraining at scale produces representations that are both compact and highly discriminative.

\paragraph{Feature redundancy.} DINOv2's Tiny ImageNet results reveal a surprising pattern: PCA at 199D (79.83\%) exceeds Full at 384D (78.25\%). This means that \emph{even unsupervised} dimensionality reduction improves accuracy, indicating significant task-irrelevant variance in the self-supervised features. All nine reduction methods improve over Full on DINOv2/Tiny ImageNet.

\paragraph{Transformer robustness to reduction.} On CUB-200, the accuracy gap between Full and the best reduction method is 1.21--1.43\% for transformers versus 1.23--3.50\% for CNNs. Transformer features encode discriminative information in a more distributed, lower-rank manner, making them more resilient to linear projection.

\paragraph{LDA benefit generalizes.} On coarse-grained tasks, LDA improves over Full for both CNN backbones (+0.23\% to +4.46\%) and transformer backbones (+0.08\% to +1.65\%). The benefit is not specific to convolutional feature spaces---the Fisher criterion applies equally to CLS token embeddings.

\subsection{Why Does LDA Help?}

The consistent benefit of LDA over full features on coarse-grained tasks warrants explanation. We identify three complementary mechanisms:

\paragraph{Implicit regularization.} By projecting from $D$ to $C-1$ dimensions, LDA eliminates feature dimensions that contribute to overfitting in the downstream classifier. LDA acts as a structured alternative to stronger $\ell_2$ penalties, one that uses label information to select which dimensions to keep.

\paragraph{Removal of task-irrelevant variance.} ImageNet features encode information about 1,000 categories. For a 100-class target task, much of this information is irrelevant. PCA retains high-variance directions regardless of task relevance; LDA explicitly retains directions that separate the target classes.

\paragraph{Improved conditioning for the classifier.} High-dimensional feature spaces often contain near-collinear feature groups, which slow convergence of iterative solvers like LBFGS. LDA produces orthogonal, decorrelated features that are better conditioned, leading to faster and more stable classifier training.

The magnitude of LDA's gain correlates with how much the classifier benefits from regularization. On compact backbones (ResNet-18 at 512D, MobileNetV3 at 576D), where the full-feature logistic regression is least constrained relative to the task complexity, LDA provides the greatest lift (+2.8 to +4.5\%). On high-capacity backbones (ResNet-50 at 2048D, ViT-B/16 at 768D), the full features already have ample capacity and LDA's gain is modest (+0.08 to +0.85\%).

\subsection{Practitioner Guidelines}
\label{sec:guidelines}

Based on our comprehensive evaluation across six backbones, three datasets, and 180 configurations, we offer the following concrete recommendations:

\begin{enumerate}[leftmargin=*]
    \item \textbf{Check your task type first.} On \emph{coarse-grained} classification (object categories, scene types), LDA reliably improves accuracy while reducing dimensionality by 48--87\%. On \emph{fine-grained} tasks (species, models, varieties), skip dimensionality reduction and classify full features directly.

    \item \textbf{Set $d = C - 1$.} This is the maximum number of LDA components and consistently the best choice (Section~\ref{sec:components}). Lower values sacrifice accuracy without meaningful speed gains.

    \item \textbf{Skip PCA+LDA.} Two-stage reduction provides no consistent benefit over direct LDA and adds unnecessary complexity.

    \item \textbf{Skip LFDA and NCA.} Neither improves on LDA for frozen features, and both are substantially slower.

    \item \textbf{Consider DSB for coarse-grained tasks if marginal accuracy matters.} DSB wins outright on 6 of 18 configs at 2--3$\times$ LDA's cost. For most applications, the gain does not justify the cost.

    \item \textbf{Use R-LDA when $D \gg C$ and the training set is moderate.} Regularization helps on high-dimensional backbones (ResNet-50, EfficientNet) with many classes.

    \item \textbf{If the training set has $< 50$ samples per class, use full features.} LDA's scatter estimates need sufficient data to be reliable.

    \item \textbf{Standardize after projection.} Feature scaling before the classifier is essential for fair and stable results regardless of the reduction method.

    \item \textbf{Expect the largest gains from compact backbones.} ResNet-18 and MobileNetV3 benefit most from LDA (+2.8 to +4.5\%), while ViT-B/16 and ResNet-50 show smaller gains.
\end{enumerate}

\section{Conclusion}
\label{sec:conclusion}

We have presented the most comprehensive empirical study to date of dimensionality reduction applied to frozen pretrained features for image classification, evaluating ten methods across six backbones (four CNNs and two vision transformers) and three datasets spanning coarse-grained and fine-grained recognition. The study comprises 180 systematically controlled experiments.

The central finding is two-fold. On \emph{coarse-grained} tasks (CIFAR-100, Tiny ImageNet), Linear Discriminant Analysis---a technique dating to 1936---improves classification accuracy over full features in 11 of 12 backbone--dataset configurations, with gains of up to 4.5 percentage points that are statistically significant at $p < 0.001$. It simultaneously reduces dimensionality by 48--87\% and accelerates downstream classifier training by 7--8$\times$ on average. On \emph{fine-grained} recognition (CUB-200), however, full features win in all six configurations, establishing a clear boundary condition: LDA's mean-based projection discards the subtle local cues that distinguish visually similar subcategories.

This boundary condition is itself a contribution. Prior work has evaluated LDA-based reduction only on coarse-grained tasks, leaving practitioners without guidance for fine-grained domains. Our results provide a simple decision rule: \emph{check the task granularity before choosing a pipeline.}

Among more sophisticated alternatives, only DSB---a boosting-inspired reweighting of the scatter matrices---provides a consistent improvement over plain LDA on coarse-grained tasks, winning 6 of 12 configurations at 2--3$\times$ the computational cost. LFDA, NCA, and two-stage PCA+LDA all fail to improve on plain LDA for frozen features. These negative results are equally valuable, sparing practitioners from adopting more complex methods without justification.

Our inclusion of vision transformers (ViT-B/16 and DINOv2) extends the study's relevance beyond CNN pipelines. LDA's benefit generalizes to CLS-token embeddings, and DINOv2's self-supervised features prove exceptionally compact: at only 384 dimensions, DINOv2 achieves the highest accuracy on both CIFAR-100 (82.41\%) and CUB-200 (87.73\%), with LDA still providing a +1.65\% gain on CIFAR-100.

\paragraph{Limitations.} Our study evaluates three datasets with 100--200 classes. Whether LDA's advantage persists at the scale of thousands of classes (ImageNet-1K, iNaturalist) remains open, as the $C-1$ dimensional projection then retains a much larger fraction of the original space. We use a single classifier paradigm (logistic regression); preliminary MLP experiments suggest that non-linear classifiers may internalize the regularization that LDA provides, reducing its benefit. Our CUB-200 results characterize the fine-grained boundary but do not resolve it---developing reduction methods that preserve local discriminative structure for fine-grained tasks is an important open problem.

\paragraph{Future work.} Several directions follow naturally. First, investigating class-conditional or local variants of LDA (e.g., sub-class discriminant analysis) that may recover performance on fine-grained tasks. Second, extending to large-scale datasets with thousands of classes where the $C-1$ projection retains significant dimensionality. Third, exploring LDA as a differentiable layer that can be trained end-to-end within frozen-backbone pipelines, bridging classical and deep metric learning. Finally, evaluating the interaction between dimensionality reduction and non-linear classifiers more thoroughly, given preliminary evidence that MLPs diminish LDA's advantage.

\paragraph{Reproducibility.} All code, data pipelines, experiment configurations, and raw results (180 CSV entries) are publicly available at \url{https://github.com/IndarKarhana/lda-image-classification}. The repository includes feature extraction scripts for all six backbones, all ten reduction methods with fixed random seeds, and scripts to regenerate every figure and table in this paper.

\bibliographystyle{IEEEtran}
\bibliography{references}

\end{document}